\documentclass[a4paper, 10pt]{article}
\pdfoutput=1
\usepackage{graphicx}
\usepackage{caption}
\usepackage{authblk}
\usepackage{subcaption}
\usepackage{amsfonts} 
\usepackage{amsmath} 
\usepackage[utf8]{inputenc}
\usepackage[english]{babel}
\usepackage{longtable}
\usepackage{booktabs}
\usepackage[margin=1in]{geometry}
\usepackage{multirow,tabularx}
\usepackage{dirtytalk}
\usepackage{fancyvrb}
\usepackage{fvextra}
\usepackage{csquotes}
\usepackage{verse}
\usepackage{multirow}
\usepackage{minted}
\usepackage{hyperref}
\usepackage[dvipsnames]{xcolor}
\definecolor{secondarycolor}{HTML}{428251}
\usepackage{xpatch}
\usepackage{soul}
\usepackage{tipa}
\usepackage[style=authoryear-comp, backend=biber, bibencoding=utf8,
    defernumbers=true, maxnames=3, maxbibnames=99, language=english, dashed=false
]{biblatex}
\hypersetup{
    colorlinks, 
    citecolor=secondarycolor, 
    linkcolor=secondarycolor, 
    urlcolor=secondarycolor,
}
\xpatchbibmacro{cite}
  {\usebibmacro{cite:label}%
   \setunit{\printdelim{nonameyeardelim}}%
   \usebibmacro{cite:labeldate+extradate}}
  {\printtext[bibhyperref]{%
     \DeclareFieldAlias{bibhyperref}{default}%
     \usebibmacro{cite:label}%
     \setunit{\printdelim{nonameyeardelim}}%
     \usebibmacro{cite:labeldate+extradate}}}
  {}
  {\PackageWarning{biblatex-patch}
     {Failed to patch cite bibmacro}}

\xpatchbibmacro{cite}
  {\printnames{labelname}%
   \setunit{\printdelim{nameyeardelim}}%
   \usebibmacro{cite:labeldate+extradate}}
  {\printtext[bibhyperref]{%
     \DeclareFieldAlias{bibhyperref}{default}%
     \printnames{labelname}%
     \setunit{\printdelim{nameyeardelim}}%
     \usebibmacro{cite:labeldate+extradate}}}
  {}
  {\PackageWarning{biblatex-patch}
     {Failed to patch cite bibmacro}}

\begin{document}

%===============================================================
% Title

\title{Training Data Size Sensitivity in Unsupervised Rhyme Recognition}
%\author[1]{author}
\author[1]{Petr Plecháč}
\author[1,2]{Artjoms Šeļa}
\author[1,3]{Silvie Cinková}
\author[4]{Mirella De Sisto}
\author[5]{Lara Nugues}
\author[6]{Neža Kočnik}
\author[1]{Antonina Martynenko}
\author[7]{Ben Nagy}
\author[8]{Luca Giovannini}
\author[1]{Robert Kolár}

%\affil[1]{affiliation}
\affil[1]{Institute of Czech Literature, Czech Academy of Sciences, Czechia}
\affil[2]{University of Tartu, Estonia}
\affil[3]{Charles University, Czechia}
\affil[4]{Tilburg University, Netherlands}
\affil[5]{University of Basel, Switzerland}
\affil[6]{University of Ljubljana, Slovenia}
\affil[7]{Institute of Polish Language, Polish Academy of Sciences, Poland}
\affil[8]{University of Potsdam, Germany}
\maketitle          

%===============================================================
% Abstract
\begin{abstract} 
\noindent Rhyme is deceptively intuitive: what is or is not a rhyme is constructed historically, scholars struggle with rhyme classification, and people disagree on whether two words are rhymed or not. This complicates automated rhymed recognition and evaluation, especially in multilingual context. This article investigates how much training data is needed for reliable unsupervised rhyme recognition using \texttt{RhymeTagger}, a language-independent tool that identifies rhymes based on repeating patterns in poetry corpora. We evaluate its performance across seven languages (Czech, German, English, French, Italian, Russian, and Slovene), examining how training size and language differences affect accuracy. To set a realistic performance benchmark, we assess inter-annotator agreement on a manually annotated subset of poems and analyze factors contributing to disagreement in expert annotations: phonetic similarity between rhyming words and their distance from each other in a poem. We also compare \texttt{RhymeTagger} to three large language models using a one-shot learning strategy. Our findings show that, once provided with sufficient training data, \texttt{RhymeTagger} consistently outperforms human agreement, while LLMs lacking phonetic representation significantly struggle with the task.
\end{abstract}

%===============================================================
% Introduction
\section{Introduction}

Rhyme is deceptively intuitive: what is or is not a rhyme is constructed and negotiated historically \parencite{houston_rhymefindr_2025} in a given literary tradition, scholars struggle with rhyme typology and classification \parencite{nagy_rhyme_2022}, and people can often disagree on whether two words are rhymed or not. This complicates automated rhymed recognition and evaluation, especially in multilingual context, with different languages having different morphological capacity for rhyming and setting distinct limitations for recognition algorithms.  In this paper we investigate how much training data is needed for reliable unsupervised rhyme recognition using \texttt{RhymeTagger} (\url{https://github.com/versotym/rhymetagger}), a language-independent tool that identifies rhymes based on repeating patterns in poetry corpora. 

\texttt{RhymeTagger} works based on a simple assumption: since the possibilities for rhyming are finite in a language, a certain portion of rhyme pairs will inevitably reappear in a sufficiently large poetry corpus. To identify such pairs, the $T$-score \parencite{church1990}, a common collocation extraction method, is used to detect word pairs that co-occur at the ends of lines significantly more often than expected by chance. The extracted pairs serve as a training set to learn the probabilities of sound combinations forming a rhyme, which are then used to annotate the entire corpus (see \cite{plechac2018} for details).

A question remains, however: what does 'sufficiently large' mean? How many lines or poems are needed to ensure enough repeating rhymes for reliable rhyme recognition? In this article, we aim to address this by evaluating \texttt{RhymeTagger}'s sensitivity to training data size across seven European languages, namely Czech (\textit{cs}), German (\textit{de}), English (\textit{en}), French (\textit{fr}), Italian (\textit{it}),\footnote{Due to the excessively broad time span of the Italian corpus (13th to 20th centuries), all samples in this study were drawn exclusively from the subcorpus of authors born after 1700.} Russian (\textit{ru}), and Slovene (\textit{sl}). Beside this, we also compare \texttt{RhymeTagger}'s performance to three large language models (LLMs), namely \texttt{GPT4-o}, \texttt{Claude 3.7 Sonnet}, and \texttt{DeepSeek-V3}. We discuss disagreement in expert human annotations and identify rhyme features that contribute to it. We show that \texttt{RhymeTagger} can outperform human agreement, while LLMs still significantly struggle with tasks that require phonetic representation of language.

%===============================================================
% Inter-Annotator Agreement
\section{Inter-Annotator Agreement}

Studies evaluating rhyme recognition typically rely on gold standards produced by a single annotator (e.g., \cite{reddy2011, plechac2018}). Rhyme, however, is not a strictly defined feature, and different annotators may not always agree. To address this, we employ multiple human-produced annotations and assess inter-annotator agreement (IAA). For evaluation purposes, we define a rhyme as a link between any two lines that belong to the same rhyme chain. A typical octave of a Petrarchan sonnet---eight lines with only two chains of rhymes ($a_1 b_1 b_2 a_2 a_3 b_3 b_4 a_4$)---would thus give $\binom{4}{2} + \binom{4}{2} = 12$ rhymes, namely: 

\begin{table}[h!]
\centering
\begin{tabular}{llllll}
$a_1:a_2$, & $a_1:a_3$, & $a_1:a_4$, & $a_2:a_3$, & $a_2:a_4$, & $a_3:a_4$, \\
$b_1:b_2$, & $b_1:b_3$, & $b_1:b_4$, & $b_2:b_3$, & $b_2:b_4$, & $b_3:b_4$ \\
\end{tabular}
\end{table}

\noindent We drew a random sample of 100 poems for each language from the PoeTree collection \parencite{poetree2024}, with each sample ranging from 1900 to 2400 lines in total.
Each sample was processed by two different expert annotators using an ad hoc web interface built for this task. Although Cohen's $\kappa$ \parencite{cohen1960} is considered a general standard for measuring IAA (and has already been used for evaluation of rhyme recognition in \cite{haider2018}), it requires a set of negative cases. This is---similarly to named entity recognition for example---quite tricky with rhymes. If one were to consider all combinations of line pairs that are not marked as rhyming, the $\kappa$ score would be greatly overestimated, since the number of negative cases would vastly outnumber the positive ones. As a simple example, a poem consisting of four quatrains \textit{abab cdcd efef ghgh} would yield 8 positive cases and 112 negative ones. For this reason we chose to measure IAA using $F_1$-score instead, as in e.g., \cite{hripcsak2005}. 

As \autoref{tab1} shows, annotator $F_1$-scores vary across languages between 0.86 and 0.96, with a median value of 0.88 and \textit{ru} being the top outlier. While these may seem to be rather modest values for a seemingly straightforward feature like rhyme, two complicating factors appear to be at play. First, annotators seem to differ as how many intervening lines they are willing to accept for a rhyming pair. While, for instance, in \textit{fr} annotator 1 does not hesitate to annotate rhymes---with no other member of the rhyme chain between them---10 or even 20 lines apart, annotator 2 never goes beyond 6 lines. This factor may also explain the notably high agreement in \textit{ru}, where the most distant rhyme pair was only 8 lines apart, and neither annotator otherwise marked rhymes beyond a 4-line distance. The second factor where annotators seem to differ is their tolerance of imperfect rhymes (e.g., memory \textipa{["mEm@\*ri]} : amity \textipa{["\ae mIti]} (\textit{en}) or Verlangen \textipa{[fEK"laN@n]} : Flammen \textipa{["flam@n]} (\textit{de})), or, especially in languages with archaic orthographies, rhymes based on historical pronunciations that may have shifted over time.

\begin{table}[t]
    \centering
    \begin{tabular}{l c}
        %\toprule
        sample & $F_1$-score \\
        \midrule
        cs & 0.90 \\
        de & 0.88 \\
        en & 0.88 \\
        fr & 0.86 \\
        it & 0.94 \\
        ru & 0.96 \\
        sl & 0.88 \\
        \bottomrule
    \end{tabular}
    \captionof{table}{Inter-annotator agreements in individual language samples}
    \label{tab1}
\end{table}

To test this effect, we selected all rhyme pairs that were marked by at least one annotator and occur consecutively within a given rhyme chain (i.e., their indices differ by one). For each language, we represent each such pair using three variables:

\begin{enumerate}
    \item \texttt{agreement} (boolean): Whether rhyme pair was annotated by both or just one annotator.
    \item \texttt{line\_distance} (integer): Number of lines between the rhyme pair.
    \item \texttt{phon\_distance} (float): We transcribe each line into IPA by means of \texttt{eSpeak-NG} (\citeyear{espeak2023}), and trim everything before the nucleus of the last stressed syllable (a common starting point of sounds match in rhyme). For each pair of lines that was annotated as rhyming at least by one of the annotators we measure the edit distance between the articulatory features of the respective phonemes using \texttt{PanPhon} \parencite{panphon2016}. 
\end{enumerate}

\noindent Next we fit a binary logistic regression using a Bayesian mixed-effects model, modeling the probability of annotator agreement with \texttt{line\_dist} and \texttt{phon\_dist} as predictors (\autoref{tab2}). The model was stratified by corpus using partial pooling: this estimates the global effect of \texttt{line\_dist} and \texttt{phon\_dist} while also allowing differing effect strengths for the various causal factors that exist per-corpus. To name just a few, some pairs of annotators will have more similar aesthetic tastes, and some languages, and the poetry written in them, may have more (or less) clear-cut rhymes on average. Although it is not possible to separate these causes, they can be handled together. The results confirm that \texttt{line\_dist} and \texttt{phon\_dist} have a negative effect on the annotator agreement (weaker rhymes, and more distant rhymes, both make it more likely that annotators will disagree), with the effect of \texttt{line\_dist} being the stronger of the two (with a credible range of $[-0.943,-0.750]$ vs $[-0.544,-0.424]$, log-likelihood). In terms of the baseline likelihood of agreement among the corpora, the \textit{cs} and \textit{ru} corpus annotations were significantly more likely to agree than the rest. Note that in \autoref{tab1} we see the $F_1$ score for \textit{ru} is markedly higher than \textit{cs}---the model confirms that this is probably not due to the overall corpus effect (the effect sizes in \autoref{tab2} are similar) but instead to the fact that both \textit{ru} annotators were disinclined to tag distant rhymes, as discussed above.

\begin{table}[t]
\centering
\begin{tabular}{lrrrr}
% \toprule
 & Mean & SD & HDI 3\% & HDI 97\% \\
\midrule
phon\_dist & -0.487 & 0.032 & -0.544 & -0.424 \\
line\_dist & -0.842 & 0.052 & -0.943 & -0.750 \\
cs & 3.356 & 0.180 & 3.027 & 3.690 \\
de & 2.237 & 0.116 & 2.026 & 2.459 \\
en & 2.338 & 0.109 & 2.126 & 2.534 \\
fr & 2.630 & 0.123 & 2.402 & 2.865 \\
it & 2.935 & 0.138 & 2.677 & 3.188 \\
ru & 3.309 & 0.182 & 2.985 & 3.672 \\
sl & 2.803 & 0.146 & 2.535 & 3.078 \\
\bottomrule
\end{tabular}
\captionof{table}{Logistic regression results by corpus. The mean effect sizes are shown, along with the 94\% High Density Interval that represents the `credible range' of the estimated effect.}
\label{tab2}
\end{table}

Taken together, the results show that across languages there is no unanimous agreement among human annotators---whether due to the factors discussed above or other variables not accounted for. Therefore, the appropriate benchmark for evaluating machine-driven rhyme recognition should be the level of agreement among humans, rather than attempting to overfit to the idiosyncrasies of a single annotator.

%===============================================================
% Evaluation of Unsupervised Learning
\section{Evaluation of Unsupervised Learning}

To evaluate \texttt{RhymeTagger}'s sensitivity to training data size we proceed as follows: 
For each of the seven languages, we randomly sampled the respective PoeTree corpus 100 times. Each sample consisted of approximately 1,000 lines and included only complete poems. Poems were sampled with replacement across samples, allowing a given poem to appear in more than one sample, but never more than once within any single sample. 
These samples are then used to train the \texttt{RhymeTagger} models. This procedure is repeated for samples of increasing size: \{2k, 3k, 4k, ..., 10k, 20k, 30k, ..., 100k, 200k, 300k, ... 1M\} (if the corpus size permits). As a result, up to 1,800 different models are trained per language.

\begin{figure}[t]
    \centering
    \includegraphics[width=0.6\textwidth]{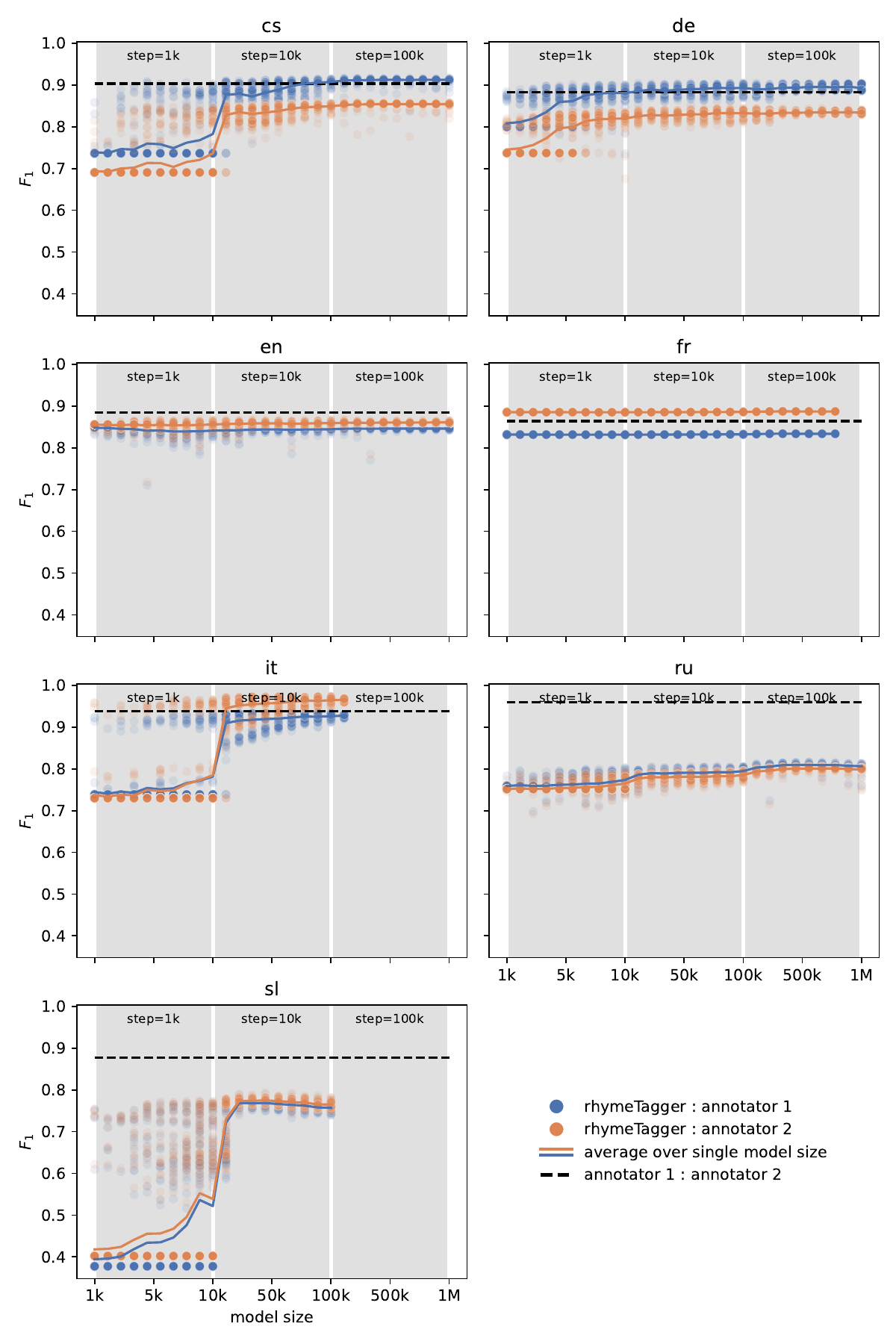}
    \caption{$F_1$-scores achieved by \texttt{RhymeTagger} against human annotators. 100 models per single model size}
    \label{fig3}
\end{figure}

\autoref{fig3} shows $F_1$-scores of these models achieved on samples processed by human annotators.
In \textit{en} and \textit{fr}, the results remain consistent across the entire range---from 1k-line models to those trained on 1M lines. We assume this is due to the predominance of perfect rhyme matches in these languages, which require little learning from the training data for \texttt{rhymeTagger} to recognize. In other languages, by contrast, the expected trend emerges: $F_1$-scores generally increase with model size up to a certain point, after which further growth brings no relevant improvement and the scores stabilize. However, the point of stabilization varies by language: in \textit{de}, it occurs around 8,000 lines; in \textit{it} around 20,000 lines, in \textit{sl}, around 30,000 lines; in \textit{cs}, approximately 60,000; and in \textit{ru}, not before 200,000 lines.

These observations are---to a certain extent---consistent with the hypothesis raised in \cite{plechac2018}, namely that one of the key factors affecting the amount of data needed is the degree of language inflection. From low-inflection languages (\textit{en}) through moderately inflected ones (\textit{de}, \textit{fr}, \textit{it}) to highly inflected Slavic languages (\textit{cs}, \textit{ru}, \textit{sl}), the amount of training data needed tends to increase. The reasoning is straightforward: [since the morphological suffixes are shared between many words, there are many more possible rhyme pairs, and so the chance of observing any given pair is lower---and thus more training data is needed] the more suffixes a language allows, the more diverse its rhyming possibilities become, and the less likely it is that rhyme pairs will reappear multiple times within a corpus of a given size, hence more training data is needed.

Once the point of stabilization is reached, \texttt{rhymeTagger} outperforms IAA in four out of seven languages (\textit{cs}, \textit{de}, \textit{fr}, and \textit{it}): that is, it aligns on average more closely with one human annotator than that annotator does with another human.
In \textit{en} its $F_1$-scores are only slightly lower than the IAA, while in \textit{ru} and \textit{sl} they are considerably worse. In the case of \textit{ru}---besides the exceptionally high IAA values themselves---this seems to be due to systematic errors in phonetic transcription: the misplacement of stresses in multi-syllable words under metrical constraints and failure to treat orthographic 'e' as /o/ after palatalized consonants. In \textit{sl}, this seems to be primarily due to both annotators' high tolerance towards distant rhymes, rhymes that were far beyond the 7-line frame \texttt{rhymeTagger} was configured to look for.

%===============================================================
% LLMs
\section{Evaluation of One-Shot Learning}

Having estimated the accuracy of the unsupervised learning, we next compare its performance to that of large language models (LLMs) using a one-shot learning strategy. Three LLMs were selected for this task: \texttt{GPT-4o}, \texttt{Claude 3.7 Sonnet}, and \texttt{DeepSeek-V3}.

Each model was queried via REST API, receiving a single text from the human-annotated test data per prompt. The task was to identify all rhymes in the given text and return a list of rhyming words in JSON format.  Each prompt included an instruction block and an example of short poem (a limerick) along with the expected output format. The prompt was structured as follows:\\[5mm]

\begin{verbatim}
    USER: 
        You are an expert in poetry. Your task is to identify end-of-line 
        rhymes in the given text. Focus exclusively on rhymes that occur 
        at the end of each line. Ignore internal or slant rhymes unless 
        they match at the end of the line.  Return your output as a JSON 
        object containing lists of rhyming words, grouped together. If a 
        word appears in multiple rhyming lines, repeat it in the output 
        as many times as it appears.
    ASSISTANT:
        EXAMPLE:
        Text: 
        There was an Old Man with a beard,
        Who said, 'It is just as I feared!
        Two Owls and a Hen,
        Four Larks and a Wren,
        Have all built their nests in my beard!'
        {"rhymes": [["beard", "feared", "beard"], ["hen", "wren"]]}: 
\end{verbatim}

\noindent The rhyme chains returned by the models were then mapped back to the original poems and the $F_1$-scores against human annotators were measured (\autoref{tab3}). (An initial attempt to have the models return indices of rhyming lines was ultimately unsuccessful.)

\begin{table}[t]
   \centering
    \begin{tabular}{lccccccc}
        %\toprule
      \multirow{2}{*}{sample} & \multicolumn{2}{c}{GPT-4o} & \multicolumn{2}{c}{Claude 3.7 Sonnet} & \multicolumn{2}{c}{DeepSeek-V3} & \multirow{2}{*}{\textit{rhymeTagger}}\\
         & annot.\ 1 & annot.\ 2 & annot.\ 1 & annot.\ 2 & annot.\ 1 & annot.\ 2 \\
        \midrule
        cs & 0.61 & 0.57 & \textbf{0.67} & 0.59 & 0.46 & 0.48 & \textit{0.91}\\
        de & \textbf{0.68} & 0.64 & 0.53 & 0.51 & 0.57 & 0.60 & \textit{0.89} \\
        en & 0.57 & 0.59 & 0.63 & 0.64 & 0.66 & \textbf{0.67} & \textit{0.86} \\
        fr & 0.77 & 0.81 & 0.76 & \textbf{0.86} & 0.62 & 0.58 & \textit{0.89}\\
        it & \textbf{0.82} & \textbf0.79 & 0.74 & 0.75 & 0.43 & 0.41 & \textit{0.97}\\
        ru & 0.64 & 0.53 & \textbf{0.67} & \textbf{0.67} & 0.60 & 0.60  & \textit{0.81}\\
        sl & 0.31 & 0.33 & 0.52 & \textbf{0.54} & 0.41 & 0.38  & \textit{0.76}\\
        \bottomrule
    \end{tabular}
    \captionof{table}{Agreement ($F_1$-score) between LLMs (one-shot learning) and human annotators. Highest scores for each language highlighted in bold. The last column gives the value for the largest available \texttt{rhymeTagger} model (higher of the two $F_1$ averages for given model size)}
    \label{tab3}
\end{table}

The best results were achieved by \texttt{Claude} which outperformed the other models in four out of seven languages (\textit{cs}, \textit{fr}, \textit{ru}, \textit{sl}), yielding the highest $F_1$-score against either of the annotators. \texttt{GPT} performed best in \textit{de} and \textit{it}. Its performance in the remaining languages trails slightly behind \texttt{Claude}, though the difference is generally modest---except for \textit{sl} (for which it is likely under-resourced). \texttt{DeepSeek} tended to lag behind the other two models, with the exception of \textit{en} where it---surprisingly---achieves the best performance.

However, none of the models in any of the languages achieves scores comparable to those of \texttt{rhymeTagger}. \texttt{Claude} is only getting close in case of \textit{fr} where it yields the value of IAA. In other languages the performance of all LLMs fall well short of that achieved by \texttt{rhymeTagger}.

The reason robust LLMs are vastly outperformed by a relatively straightforward machine learning algorithm is likely due to their lack of any phonetic representation. This is quite obvious from the type of errors they tend to produce. LLMs often miss rhymes that are orthographically distinct (e.g., \textit{praise} : \textit{days}, \textit{drawn} : \textit{on}), while at the same time generating long chains of words that share similar graphemic endings but are phonetically far apart---e.g.,
\textit{she} : \textit{blue} : \textit{glide} : \textit{alone}. Another, albeit less frequent, type of error involves what appears to be the superimposition of frequent rhyme schemes into texts following less typical patterns. One such example is when a ballad rhyme scheme (xaxa...) was treated as paired couplets (aabb...) resulting in pairs of non-rhyming words:

\begin{verbatim}
        ["head", "stallion"],
        ["bass", "rapscallion"],
        ["freak", "poem"],
        ["sounds", "below‘ em"],
        ["art", "defend I"],
        ["spawn", "scribendi"],
        ["effects", "Hannah"],
        ["debased", "Diana"],
        ["marine", "barks on"],
        ["jar", "Clarkson"],
        ["may", "distant"],
        ["design", "consistent"]
\end{verbatim}

Ironically, LLMs that are perfectly capable of writing rhymed poetry and that strongly associate poetic texts with the presence of rhyme \parencite{walsh2024} struggle with its recognition, which highlights the asymmetry between written language capabilities of these models and their lack of phonetic sensibility.

%===============================================================
% Conclusions
\section{Conclusions}

This study set out to evaluate the sensitivity of unsupervised rhyme recognition to training data size, using the \texttt{RhymeTagger} system across seven languages. We complemented this with an assessment of IAA to establish a realistic performance ceiling for machine driven rhyme recognition, and we benchmarked \texttt{RhymeTagger} against state-of-the-art LLMs using a one-shot learning strategy.

Our findings show that while human annotators do not nearly reach an unanimous consensus---be it for variation in tolerance for imperfect rhymes or differences in how far apart rhyming lines can be---\texttt{RhymeTagger}, when provided with sufficient training data, is capable of achieving and even surpassing human-level agreement. While the amount of data needed varies by language, training on 10,000 to 50,000 lines generally yields reliable performance. In the two languages where \texttt{RhymeTagger} falls short of IAA, performance could likely be improved through better phonetic transcriptions (as in \textit{ru}) or by adjusting model parameters (as in \textit{sl}).

In contrast, LLMs showed inconsistent results and failed to match \texttt{RhymeTagger}'s performance in any language. We attribute this to their lack of explicit phonetic representations, which leads to systematic errors in rhyme detection.

%===============================================================
% Acknowledgment
\section*{Acknowledgment}

This article was supported by the Czech Science Foundation (project ga23-07727S). Data \& code available at \url{https://doi.org/10.5281/zenodo.15744239}.

%===============================================================
% Bibliography
\printbibliography

%===============================================================
\end{document}